# FathomNet: A global image database for enabling artificial intelligence in the ocean

**Kakani Katija[1,2,3,*], Eric Orenstein[1], Brian Schlining[1], Lonny Lundsten[1], Kevin Barnard[1], Giovanna Sainz[1], Oceane Boulais[4], Megan Cromwell[5], Erin Butler[6], Benjamin Woodward[6], and Katy Croff Bell[7]**

[1]Monterey Bay Aquarium Research Institute, Research and Development, Moss Landing, 95039, USA
[2]California Institute of Technology, Graduate Aerospace Laboratories, Pasadena, 91125, USA
[3]Smithsonian Institution, National Museum of Natural History, Washington DC, 37012, USA
[4]NOAA, Southeast Fisheries Science Center, Key Biscayne, 33149, USA
[5]NOAA, National Centers for Environmental Information, Stennis Space Center, 39529, USA
[6]CVision AI Inc., Research and Development, Medford, 02155, USA
[7]Ocean Discovery League, Saunderstown, 02874, USA
[*]kakani@mbari.org

## ABSTRACT

The ocean is experiencing unprecedented rapid change, and visually monitoring marine biota at the spatiotemporal scales needed for responsible stewardship is a formidable task. As baselines are sought by the research community, the volume and rate of this required data collection rapidly outpaces our abilities to process and analyze them. Recent advances in machine learning enables fast, sophisticated analysis of visual data, but have had limited success in the ocean due to lack of data standardization, insufficient formatting, and demand for large, labeled datasets. To address this need, we built FathomNet, an open-source image database that standardizes and aggregates expertly curated labeled data. FathomNet has been seeded with existing iconic and non-iconic imagery of marine animals, underwater equipment, debris, and other concepts, and allows for future contributions from distributed data sources. We demonstrate how FathomNet data can be used to train and deploy models on other institutional video to reduce annotation effort, and enable automated tracking of underwater concepts when integrated with robotic vehicles. As FathomNet continues to grow and incorporate more labeled data from the community, we can accelerate the processing of visual data to achieve a healthy and sustainable global ocean.

## Introduction

The ocean is experiencing unprecedented rapid change, and visually monitoring marine biota at the spatiotemporal scales needed for responsible stewardship is a formidable task. Monitoring a space as vast as the ocean[1] that is filled with life that we have yet to describe[2], using traditional, resource-intensive (e.g., time, person-hours, cost) sampling methodologies are limited in their ability to scale in spatiotemporal resolution and engage diverse communities[3]. However, with the advent of modern robotics[4], low-cost observation platforms, and distributed sensing[5], we are beginning to see a paradigm shift in ocean exploration and discovery. This shift is evidenced in oceanographic monitoring via satellite remote sensing of near-surface ocean conditions and the global ARGO float array, where distributed platforms and open data structures are propelling the chemical and remote sensing communities to new scales of observation[6,7]. Due to a variety of constraints, large-scale sampling of biological communities or processes below the surface waters of the ocean has largely lagged behind.

There are three common modalities for observing biology and biological processes in the ocean – acoustics, "-omics", and imaging – each with their strengths and weaknesses. Acoustics allow for observations of population- and group-scale dynamics, however individual-scale observations, especially determination of animals down to lower taxonomic groups like species, are challenging tasks[8]. The promising field of eDNA allows for identification of biological communities based on their shed DNA in the water column. While eDNA studies provide broad-scale views of biological communities with only a few discrete samples, determining the spatial source of the DNA, relating the measurements to population sizes, and the presence of confounding non-marine biological markers in samples are active areas of research that still need to be addressed[9]. Ultimately, -omics and acoustics approaches rely on visual observations for verification. Imaging, a non-extractive method for ocean observation, enables the identification of animals to the species level, elucidates community structure and spatial relationships in a variety of habitats, and reveals fine-scale behavior of animal groups[10]. However, processing visual data, particularly data with complex scenes and organisms that require expert classifications, is a resource-intensive process that cannot be scaled without significant investment, capacity building, and advances in automation[11,12].

Imaging is an increasingly common modality for sampling biological communities in a variety of environments due to the ease by which the technology can be deployed, and the number of remotely controlled and autonomous platforms that can be used[13]. Imaging has also been used for real-time, underwater vehicle navigation and control while performing difficult tasks in complex environments[14]. Moreover, imaging is an effective engagement tool to share marine life information and the issues facing the ocean with broader communities[15,16]. In short, visual data is an invaluable tool for better understanding the ocean and conveying that information broadly.

Given all the applications of marine imaging, a number of annotation tools have been developed to manage and analyze visual data. These efforts have resulted in many capable software solutions that can either be deployed locally on a computer, during field expeditions, or broadly on the worldwide web[17]. However, with the limited availability of experts and the prohibitive costs to annotate and store footage, novel methods for automated annotation of marine visual data are desperately needed. This need is motivating the development and deployment of artificial intelligence and data science tools for ocean ecology.

Artificial intelligence (AI) is a broad term that encompasses many different approaches[18], some of which have already been used to study marine systems. Statistical learning methods like random forests have been used in the plankton imaging community, achieving automated classification of microscale plants and animals with accuracies greater than 90%[11]. Unsupervised learning can be deployed with minimal data and a priori knowledge of marine environments, however these algorithms have limited application for automating the detection and classification of objects in marine imagery with sufficient granularity and detail to be used for annotation[19]. Deep learning algorithms trained on visual data where all objects have been identified (e.g., Figure 1) have improved performance of automated annotation and classification tasks to finer taxonomic levels[20,21], however this approach requires publicly available, large-scale labeled image datasets for training[22–24].

Image repositories for terrestrial applications have been available to the computer vision (CV) community for many years. ImageNet was the first labeled set based on a hierarchy, or the number of classes (or "things") in WordNet, with the long-term goal to collect 500 to 1k full-resolution images for 80k concepts, or ∼50M images[22]. In order to reach this scale, ImageNet used images scraped from Flickr, resulting in a collection of largely iconic images (e.g., centered objects in relatively uncluttered environments). Like ImageNet, Microsoft's COCO[23] used workers with Amazon's Mechanical Turk to generate labels (5k labeled instances) for images with 90 classes, resulting in 2.5M labeled instances in more than 328k images. More recently, iNat2017 is a biologically focused dataset built from 675k images of 5k species of animals that have been collected and verified by users of iNaturalist[25]. Unlike ImageNet and iNat2017, COCO was specifically built to include non-iconic views of "things" that provides imagery with contextual relationships between objects, which is especially relevant to marine environments.

Large, publicly available labeled image datasets for marine concepts primarily represent planktonic communities. Imaging at these spatial scales (∼10s-1,000s of microns) requires controlled lighting and imaging conditions that utilize darkfield, brightfield, or holographic illumination[26], and these datasets include WHOI-Plankton[27] (3M images representing 109 concepts collected by 1 imaging system) and EcoTaxa[11] (150M images collected by 10 imaging systems) among others. While the plankton imaging community has made significant progress in creating image data portals, these large datasets are primarily built for classification tasks (i.e., contains regions of interest only) and necessarily exclude larger animals and other marine animals found in midwater and benthic environments. CoralNet[28], a portal and platform for working with coral imagery, is of a similar order of magnitude to these plankton datasets but is likewise restricted to a particular type of organism. There are no equivalent datasets for macroscopic ocean organisms; image sets of higher trophic-level animals[29] are typically distributed in individual repositories that can be difficult to find for non-subject matter experts in the CV and AI communities. Thus, there is a clear need for a labeled image dataset that is representative of diverse biological communities across spatial scales in the ocean that can be readily accessed in a single, publicly available online repository (Figure 1).

## Results

In order to process imagery and video collected anywhere in the world's ocean, we need a comprehensive labeled image dataset that can scale to the global ocean (Figure 1). If we want to enlist individuals that are not subject-matter experts in marine imaging but rather computer scientists that are familiar with developing state-of-the-art algorithms for automated image analysis, we need an accessible central repository for this labeled data[22,23]. To address this need, we developed FathomNet, a distributed, publicly available labeled image dataset built on FAIR data principles[30], that uses community-recognized Darwin CORE archive data formats (Table 1)[31]. FathomNet has a Rest API and website (www.fathomnet.org; Figure 2) backed by a relational database (Figure S1) that integrates widely used community web services (e.g., World Register of Marine Species or WoRMS[32]; Figure 3).

### FathomNet database architecture

FathomNet is a SQL server database that utilizes various web services (e.g., VARS, WoRMS, MarineRegions[33]), and NATS.io message bus (Figure 3). Data can either be accessed through the FathomNet website[34] (Figure 2) or via the FathomNet Rest API. FathomNet currently utilizes either WoRMS[35] or MBARI's knowledgebase (Figure S1), and can accommodate other taxonomies that have existing APIs. Images, image URLs, and associated metadata (Table 1) can be uploaded on the FathomNet database via either the website using a CSV file and appropriate Darwin CORE archive fields (Table 1), or the Rest API. Fields such as *imaging type* and *alt concept* allow users to provide additional details about their data, including whether collected by particular platform or imaging system (e.g., ROV, Baited Remote Underwater Video[36], or Underwater Video Profiler[37]), or identifying subfeatures (e.g., head or tail) for a particular concept, respectively. Submission of data is open to anyone granted write access to the database, and data will undergo a quality control process of verification by members of the FathomNet community.

The FathomNet database seeks to aggregate marine labeled image data for more than 200k currently accepted species of *Animalia* in the WoRMS database[32] using community-based taxonomic standards[38]. Our goal is to obtain 1,000 independent observations for each species in diverse poses and imaging conditions, resulting in more than 200 million observations that will continue to grow as the number of described species increase. While we use species in *Animalia* as an initial target, the FathomNet concept tree (Figure S1) is currently based on MBARI's Video Annotation and Reference System (VARS) knowledgebase[39], and can be expanded beyond biota to include underwater instances of equipment, geological features, marine debris, etc. Additional taxonomic hierarchies[40,41] can be wired into FathomNet via respective web service API calls. By utilizing existing annotation tools and providing functionality that will enable reviewing, editing, augmenting, and verifying submitted data (Figure 2), we can aggregate existing underwater image datasets that can be added to the database.

### FathomNet dataset statistics

FathomNet is seeded with data from three sources, collectively representing 30+ years of ROV, AUV, drop camera, and stereo imaging deployments: (1) MBARI's VARS database, (2) National Geographic Society's (NGS) Deep Sea Camera System[42], and (3) the National Oceanic and Atmospheric Administration's (NOAA) ROV *Deep Discoverer*[43]. As of July 2022, FathomNet has 84,454 images, 175,873 localizations from 204 separate collections for 2,244 concepts, with additional contributions ongoing. This snapshot of the database is between ImageNet[22] and COCO[23] in terms of the numbers of categories and instances per image (Figures 4a-c). The 13 super categories in the iNat2017[25] benchmark dataset (collapsed from all 5,039 fine-grained concepts of terrestrial organisms from iNaturalist) contains many more instances per concept relative to FathomNet at the class level (Figure 4c), however FathomNet has more diversity in terms of the number of concepts and instances per image (Figure 4a,b). Relative to ImageNet, COCO, and iNat2017, the localizations in FathomNet typically occupy less of the frame and occur in diverse poses due to inclusion of iconic and non-iconic imagery (Figures 4d, 5a). Coverage, a characteristic that defines whether all objects within an image are annotated, varies depending on the environment the image was captured (e.g., benthic or midwater) and the taxonomic rank (Figure 5b). Images of midwater organisms are typically more completely annotated than those taken on the seafloor on a per frame basis, and is largely a function of the population density of benthic organisms such as deep sea corals and urchins (Figure 1).

## Representative FathomNet use cases

To demonstrate the utility of FathomNet, we present three forward-looking use cases: object detection of animals in underwater images, activity recognition (presence of objects) in underwater video, and machine learning algorithms integrated in underwater vehicle controllers[14] for automated animal tracking.

### *Benthic animal detection using NOAA footage and MBARI's training data*

One of the benefits of FathomNet will be to use the training data and models to supplement the annotation of video and imagery collected in other regions of the ocean by other researchers and imaging systems. To evaluate this functionality, we fine-tuned a RetinaNet object detection model with a ResNet backbone on 20 high-level benthic supercategories (urchin, fish, sea cucumber, anemone, sea fan, sea star, worm, sea pen, crab, glass sponge, shrimp, ray, flatfish, squat lobster, gastropod, eel, soft coral, feather star, sea spider, and stony coral) generated by combining approximately 1k taxonomic classes found in the MBARI taxonomic knowledgebase (Figure S1)[44]. We hypothesized that (1) the high-level taxonomic model would perform more robustly than one including all fine-grained classes and (2) would be extensible to other regions despite training data coming mostly from a single region (the Northeast Pacific). The benthic model[44] was then used to generate algorithm proposals on frame grabs from the NOAA CAPSTONE project[45] during an expedition to the Musician Seamounts. Experts in MBARI's Video Lab subsequently annotated and localized a subset of these frame grabs, and the data were compared with the model-generated proposals.

The model performed reasonably well when tested on images from Monterey Bay for the concept coverage analysis (Figure 6a). The row of background false negatives is an artifact of the original training data coverage. The model recorded many background false positives and negatives in the target NOAA data, sometimes overwhelming the number of correct annotations of a given class (Figure 6b). While there is substantial overlap between the data collected at the Musician Seamount and in Monterey Bay – images collected by ROVs of benthic communities – many subtle differences confounded the automated detector: slight changes in the angle of the camera relative to the sea-floor, new taxonomic classes that were not represented in the training data, and higher densities of different organisms (Figure 6c,d). The behavior of the object detector is consistent with the effects of distribution shift, where the statistics of the target data change relative to the training data[46]. Indeed, the density and morphology of the corals represented by the sea fan supercategory along the Musician Seamount presented challenges to the detector (Figure 6c,d). The output of the MBARI model on the NOAA data is not reliable on its own, but substantially eases the task of human annotators looking to begin annotation efforts on raw data.

### *Midwater transect activity detection using NOAA footage and MBARI's training data*

An activity recognition (AR) routine was deployed using a midwater object detector[47]: a *RetinaNet* model with a *ResNet50*[48] backbone pre-trained on *ImageNet*[22]. Fine-tuning of parameters was done with in situ, underwater imagery from FathomNet and VARS of 17 target midwater animal classes. For the purpose of activity recognition, the model[47] was used as a binary object detector, and results were smoothed to identify segments of video for a human operator to review and annotate. This is a relevant use case for reviewing footage after it has been collected (e.g., post-processing of ROV expedition video), as it can be tedious and costly for an individual to watch many hours of video looking for only a few events. Presenting annotators with segments of video that only contain animals that need annotation can decrease the amount of time it takes to thoroughly annotate video footage[49].

The AR routine was applied to video data collected by NOAA's ROV *Deep Discoverer*, during an expedition to the Musician Seamounts as part of the NOAA CAPSTONE project[45]. An MBARI researcher manually annotated one hour of video with segments of interest, and these segments were compared against the AR model-generated segments (Figure 7, left). Activity detection signals were filtered by convolving with a 10 second window to fill any gaps between object detections. Intersection over union (IOU), which takes the temporal overlap of AR routine outputs and divides by total activity duration that occurs in either method, was used to quantitatively compare approaches. The IOU over all annotated videos was 43% and event-level recall (accounting for the imperfect temporal registration between activities) was 63%. A review of these results found that the object detector did not perform well when the object was blurred (either due to motion or unfocused optics; Figure 7, bottom-right) or when the animal was small relative to total field of view (animal $\sim 50$ pixels in length); in some cases (e.g., animal lacked sufficient contrast to the background) the algorithm correctly identified activities that the human annotator missed (Figure 7, top-right). Although the AR routine presented here requires further refinement (Video S1), 19% of the footage contained animals or objects of interest, and presenting annotators with video clips of interest would similarly reduce annotator effort.

### *Machine learning-integrated tracking (ML-Tracking) of animals for vehicle navigation and control*

In order to observe animal behavior in the ocean, observational platforms are required to non-invasively execute targeted sampling and maintain a persistent presence to track animals over a period of time. Vision-based underwater vehicle tracking has seen renewed interest due to developments in modern computer vision and machine learning[50–53], and methods like tracking-by-detection and integrating neural networks have shown promise for robust tracking[54]. For applications where longer duration, 24+ hr-long deployments are required (e.g., studying critical behaviors of animals particularly in the ocean's midwaters[55]), these modern techniques are needed.

Katija et al.[14] developed a Machine Learning-integrated Tracking (ML-Tracking) algorithm that incorporates a FathomNet-trained multi-class *RetinaNet*[56] detection model[47], 3D stereo tracker subroutines, and a supervisor module that sends commands to a vehicle controller. ML-Tracking algorithms were demonstrated in midwater using ROV *MiniROV* in the Monterey Bay National Marine Sanctuary. Nearly 50 hours of stereo and high-definition footage were collected by the authors to evaluate the ML-Tracking algorithm[14]. The longest continuous tracking trial involved a gelatinous animal – a siphonophore *Lychnagalma* sp. – for a duration of 18,987 s or 5.27 hrs. Applying ML-Tracking to the long-duration observation, the percentage of time the vehicle controller received 3D position information from the ML-Tracking algorithm (binned at 1 s intervals) was 100%[14].

## Discussion

FathomNet is built to aggregate data from disparate sources and accelerate discovery by giving the community access to more data and expertise. Fundamentally, the system relies on fine-grained, taxonomically correct annotations, which is a difficult standard to meet via crowd-sourcing[57]. The initial release of FathomNet is thus built on marine domain experts' annotation efforts; in contrast, ImageNet and COCO are built from images scraped from publicly available internet repositories and annotated by crowd-workers. Based on our estimates (see Methods; Table S1), the ImageNet 2012[58] and COCO 2014[23] releases cost ∼\$2464k and ∼\$85k respectively, excluding the cost of image generation and other infrastructure. In contrast, MBARI's seed contribution to FathomNet cost ∼\$165k to annotate based on estimates from MBARI's Video Lab, where experienced annotators label midwater and benthic images at \$80 an hour, or ∼\$1 and ∼\$3 per image depending on the complexity of the environment. At that same hourly rate, the 2012 ImageNet and 2014 COCO dataset release would have cost ∼\$6M and ∼\$2.1M to annotate, respectively. Importantly, FathomNet currently draws largely from the MBARI VARS database, which is comprised of 6,190 ROV dives that represents ∼\$144M worth of ship time and excludes the instrument and platform costs used to obtain that data. Including these additional costs underscores the current and future value of FathomNet, especially to groups in the ocean community that are early in their data collection process.

For FathomNet to reach its intended goals, significant community engagement, high-quality contributions across a wide range of groups and individuals, and broad utilization of the database will be needed. FathomNet can be leveraged to develop fully and semi-automated workflows that assist, but not replace, the annotation efforts of expert taxonomists and ecologists. This vision is very much in accord with decades-old aspirations of the marine biological community[59], and recent findings that even state-of-the-art machine learning systems suffer from biases[24] that can be mitigated with human intervention[60]. In addition to the database itself, an ecosystem of services with FathomNet tutorials,[61,62] code,[63] and trained models have been created to sustain and grow the user community. The FathomNet website[34] contains community features that allow for select members to review submissions and verify data contributions. A publicly accessible FathomNet Model Zoo (FMZ)[64] contains FathomNet-trained machine learning models contributed by community members to be freely used on other visual data collected in various marine regions with a number of platforms containing many concepts. As FathomNet grows to include additional concepts and imagery, we envision intellectual activities around the dataset similar to ImageNet and Kaggle-style competitions, where baseline datasets and annual challenges could be leveraged to develop state-of-the-art algorithms for future deployment[58,65].

FathomNet has the potential to increase the rate of image and video data analysis by human observers, contribute to generation of labeled data, and create smarter algorithms deployed on robotic vehicles to enable targeted sampling and persistent observations of marine animals. Benthic and midwater machine learning models trained on FathomNet data derived solely from MBARI[44,47] were used to create bounding box proposals and activity-detected video segments (Figures 6, 7) from visual data collected by other underwater platforms and institutions (e.g., NOAA's ROV *Deep Discoverer*). This resulted in a reduction of human annotator effort by 81% in one use case, and demonstrated the potential applicability of high-level taxonomic models on other institutional data. Additionally, demonstrations using FathomNet-trained algorithms integrated with underwater vehicle controllers[14] provide optimism for future AI-enabled missions to fully automate targeted sampling of marine objects, persistent observations of animals, and result in less-invasive sampling of valuable resources in the ocean.

For the foreseeable future, there is not going to be a single general-use classifier or detector for all ocean image data. There simply is not enough expert annotated data and too many undiscovered organisms to train such a system. Moreover, there is substantial evidence that contemporary models trained on benchmark datasets are not robust to natural distribution shifts at either the pixel or population level[46,66], an observation consistent with our experiments on out-of-distribution target datasets. While the MBARI and NOAA datasets are similar, differences in the imaging systems, deployment mode, and environment were enough to challenge the model. While such a model may not be deployable out-of-the-box, it could substantially alleviate the human cost of working with a new dataset, or be used to assess performance and suggest avenues for strategic retraining. We hope that FMZ facilitates such machine learning experiments, and enables more efficient workflows for all manner of organizations and groups. We believe this approach to open science for ocean image data has tremendous potential to accelerate the field as has been demonstrated in other areas[24,67].

While growth of the FathomNet database has inherent value for the research community, the potential for public engagement and its impact on education and conservation is equally important. The imagery contained within the database can be integrated into digital video data repositories and workflows using the Rest API, and could enable a global ocean life guide that can support research and education initiatives. Social media campaigns for community engagement, similar to iNaturalist, and eBird[67,68] could have similar, far-reaching outcomes for FathomNet, providing a mechanism for aggregating and leveraging taxonomic knowledge shared by the community. By making FathomNet publicly accessible, we can also invite ocean enthusiasts to solve challenges related to ocean visual data at temporal and spatial scales that are impossible without widespread engagement coupled with semi-automation. Incorporating human-AI interaction research with video gaming concepts and award structures could enable widespread participation similar to other gaming platforms[69], enabling direct science contributions by game players[70]. Through FathomNet and its community of users, we can create an ecosystem around marine visual data that can realize a more inclusive, equitable, and diverse vision of ocean exploration and discovery.

# Methods

## FathomNet seed data sources and augmentation tools

FathomNet has been built to accommodate data contributions from a wide range of sources. The database has been initially seeded with a subset of curated imagery and metadata from the Monterey Bay Aquarium Research Institute (MBARI), National Geographic Society (NGS), and the National Oceanic and Atmospheric Administration (NOAA). Together, these data repositories represent more than 30 years of underwater visual data collected by a variety of imaging technologies and platforms around the world. To be sure, the data currently contained within FathomNet does not include the entirety of these databases, and future efforts will involve further augmenting image data from these and other resources.

### *MBARI's Video Annotation and Reference System*

Beginning in 1988, MBARI has collected and curated underwater imagery and video footage through their Video Annotation and Reference System (VARS[39]). This video library contains detailed footage of the biological, geological, and physical environment of the Monterey Bay submarine canyon and other areas including the Pacific Northwest, Northern California, Hawaii, and the Gulf of California. Using eight different imaging systems (mostly color imagery and video, with more recent additions that include monochrome computer vision cameras[14]) deployed from four different remotely operated vehicles (ROVs *MiniROV*, *Ventana*, *Tiburon* and *Doc Ricketts*), VARS contains approximately 27,400 hours of video from 6,190 dives, and 536,000 frame grabs. These dives are split nearly evenly between observations in benthic (from the seafloor to 50 m above the seafloor) and midwater (from the upper surface of the benthic environment to the lower surface of the lighted shallower waters or $\sim$200 m) habitats. Image resolution has improved over the years from standard definition (SD; 640 x 480 pixels) to high-definition (HD; 1920 x 1080 pixels), with 4K resolutions (3840 x 2160 pixels) starting in 2021. Additional imaging systems managed within VARS, which include a low-light camera[1], the I2MAP autonomous underwater vehicle imaging payload, and DeepPIV[71], are currently excluded from data exported into FathomNet. In addition to imagery and video data, VARS synchronizes ancillary vehicle data (e.g., latitude, longitude, depth, temperature, oxygen concentration, salinity, transmittance, and vehicle altitude), and is included as image metadata for export to FathomNet.

Of the 27,400 hours of video footage, more than 88% has been annotated by video taxonomic experts in MBARI's Video Lab. Annotations within VARS are created and constrained using concepts that have been entered into the knowledge database (or knowledgebase; see Figure S1) that is approved and maintained by a knowledge administrator using community taxonomic standards (i.e., WoRMS[35]) and input from expert taxonomists outside of MBARI. To date, there are more than 7.5M annotations across 4,300 concepts within the VARS database. By leveraging these annotations and existing frame grabs, VARS data were augmented with localizations (bounding boxes) using an array of publicly available[72,73] and in-house[74–76] localization and verification tools by either supervised, unsupervised, and/or manual workflows[77]. More than 170,000 localizations across 1,185 concepts are contained in the VARS database and, due to MBARI's embargoed concepts and dives, FathomNet contains approximately 75% of this data at the time of publication.

### *NGS's benthic lander platforms and tools*

The National Geographic Society's Exploration Technology Lab has been deploying versions of its autonomous benthic lander platform (the Deep Sea Camera System, DSCS) since 2010, collecting video data from locations in all ocean basins[42]. Between 2010 and 2020, the DSCS has been deployed 594 times, collecting 1,039 hrs of video at depths ranging from 28 to 10,641 m in a variety of marine habitats (e.g., trench, abyssal plain, oceanic island, seamount, arctic, shelf, strait, coastal, and fjords). Videos from deployments have subsequently been ingested into CVision AI's cloud-based collaborative analysis platform Tator[73], where they are annotated by subject-matter experts at University of Hawaii and OceansTurn. Annotations are made using a Darwin Core-compliant protocol with standardized taxonomic nomenclature according to WoRMS[78], and adheres to the Ocean Biodiversity Information System (OBIS[79]) data standard formats for image-based marine biology[42]. At the time of publication, 49.4% of the video collected using the DSCS has been annotated. In addition to this analysis protocol, animals have also been localized using a mix of bounding box and point annotations. Due to these differences in annotation styles, 2,963 images and 3,256 annotations using bounding boxes from DSCS has been added to the FathomNet database.

### NOAA'S Office of Exploration and Research video data

The National Oceanic and Atmospheric Administration (NOAA) Office of Ocean Exploration and Research (OER) began collecting video data aboard the RV *Okeanos Explorer* (EX) in 2010, but only retained select clips due to the volume of the video data until 2016, when deck-to-deck recording began. As NOAA's first dedicated exploration vessel, all video data collected are archived and made publicly accessible from the NOAA National Centers for Environmental Information (NCEI)[80]. This specialized access is dependent upon standardized ISO 19115-2 metadata records that incorporate annotations. The dual remotely operated vehicle system, ROVs *Deep Discoverer and Seirios*[45] contains 15 cameras: 6 HD and 9 SD. Two camera streams, typically the main HD cameras on each ROV, are recorded per cruise. The current video library includes over 271 TB of data collected over 519 dives since 2016, including 39 dives with midwater transects. The data were collected during 3,938.5 hours of ROV time, 2,610 hours of bottom time, and 44 hours of midwater transects. These data cover broad spatial areas (from the Western Pacific to the Mid-Atlantic) and depth ranges (from 86 m to 5999.8 m). Ancillary vehicle data (e.g. location, depth, pressure, temperature, salinity, sound velocity, oxygen, turbidity, oxidation reduction potential, altitude, heading, main camera angle, and main camera pan angle) are included as metadata.

NOAA-OER originally crowd-sourced annotations through volunteer participating scientists, and began supporting expert taxonomists in 2015 to more thoroughly annotate collected video. In 2015, NOAA-OER and partners began the Campaign to Address Pacific Monument Science, Technology, and Ocean NEeds (CAPSTONE), which was a 3 year campaign to explore U.S. marine protected areas in the Pacific. Expert annotations generated by the Hawaii Undersea Research Laboratory[45] for this single campaign generated more than 90,000 individual annotations covering 187 dives (or 36% of the EX video collection) using VARS[39]. At the University of Dallas, Dr. Deanna Soper's undergraduate student group localized these expertly generated annotations for two cruises consisting of 37 dives (or 7% of the EX collection) from CAPSTONE, producing 8,165 annotations and 2,866 images using the Tator Annotation tool[73]. These data have formed the initial contribution of NOAA's data to FathomNet.

## Computation of FathomNet database statistics

Drawing several metrics from the popular ImageNet and COCO image databases[22,23], and additional comparisons with iNat2017[25], we can generate summary statistics and characterize the FathomNet dataset. These measures serve to benchmark FathomNet against these resources, underscore how it is different, and reveal unique challenges related to working with underwater image data.

### Aggregate statistics

Aggregate FathomNet statistics were computed from the entire database accessed via the Rest API in October 2021 (Figures 4, 5). To visualize the amount of contextual information present in an image, we estimated the number of concepts and instances as a function of the percent of the full frame they occupy (Figures 4a,b), with FathomNet data split taxonomically (denoted by $x$) to visualize how data breaks down into biologically relevant groupings. The taxonomic labels at each level of a given organism's phylogeny were back-propagated from the human annotator's label based on designations in the knowledgebase (Figure S1). If an object was not annotated down to the relevant level of the taxonomic tree (e.g., species), the next closest rank name up the tree was used (e.g., genus). The average number of instances and concepts are likewise split at taxonomic rank (Figure 4c). The percent of instances of a particular concept and how they are distributed across all images is shown in Figure 4d.

### Concept coverage

Coverage – an indication of the completeness of an image's annotations – is an important consideration for FathomNet. Coverage is quantified as average recall, and is demonstrated over 50 randomly selected images at each level of the taxonomic tree (between order and species; Figure S1) for a benthic and midwater organism, *Gersemia juliepackardae* and *Bathochordaeus mcnutti*, respectively (Figure 5a). This is akin to examining the precision of annotations as a function of synset depth in ImageNet[22]. FathomNet images with expert-generated annotations at each level of the tree, including all descendent concepts, were randomly sampled and presented to a domain expert. They then evaluated the existing annotations and added missing ones until every biological object in the image was localized. The recall was then computed for the target concept and all other objects in the frame. The false detection rate of existing annotations was negligible, and was much less than 0.1% for each concept.

***Pose variability: Iconic versus non-iconic imagery***

The data in FathomNet represents the natural variability in pose of marine animals, which includes both iconic and non-iconic views of the concept. A subject's position relative to the camera, relationship with other objects in the frame, the amount it is occluded, and the imaging background are all liable to change between frames. By computing the average image across each concept, an image class with high variability in pose (or non-iconic) will result in a blurrier, more uniformly gray image than a group of images with little pose diversity (or iconic)[22]. We computed the average image from an equivalent number of randomly sampled images across two FathomNet concepts (medusae and echinoidiae) and the closest associated synsets in ImageNet (jellyfish and starfish), which is shown in Figure 5b.

## FathomNet data usage and ecosystem

The FathomNet data use policy balances the need for distributed metadata sharing while simultaneously providing protection for data contributors wanting to maintain copyright of their valuable underwater image assets. All submitted annotation data are licensed under a Creative Commons Attribution - No Derivatives 4.0 International License. For image data that are submitted to FathomNet via a list of URLs, the original owner of those images maintains their copyright. The use of images that are directly or indirectly hosted by FathomNet are then licensed under a Creative Commons Attribution - Non Commercial - No Derivatives 4.0 International license, and all of the images may be used for training and development of machine learning algorithms for commercial, academic, non-profit, and government purposes. For all other uses of the images, users are to contact the original copyright holder, which can be tracked within the FathomNet database.

To grow the FathomNet community, we have created other resources that enable contributions from data scientists to marine scientists and ocean enthusiasts. Along with the FathomNet database, machine learning models that are trained on the image data can be posted, shared, and subsequently downloaded from the FathomNet Model Zoo (FMZ;[64]). Community members can not only contribute labeled image data, but also provide subject-matter expertise to validate submissions and augment existing labels via the web portal[34]. This is especially helpful when images do not have full coverage annotations. Finally, additional resources include code[63], blogs[61], and YouTube channel[62], that contain helpful information about FathomNet.

## Estimating and contextualizing FathomNet's value

Two of the most commonly used image databases in the computer vision community, ImageNet and COCO, are built from images scraped from publicly available internet repositories. Both ImageNet and COCO were built with crowd-sourced annotation via Amazon's Mechanical Turk (AMT) service, where workers are paid per label or image. The managers of these data repositories have not published the collection and annotation costs of their respective databases, however we can estimate these costs by comparing the published number of worker hours with compensation suggestions from AMT optimization studies. The recommended dollar values a study generating computer vision training data[81] and scientific annotations[82] are in keeping with several meta-analyses of AMT pay scales, suggesting that 90% of HIT rewards are less than \$0.10 a task and that average hourly wages are between \$3 and \$3.50 per hour[83,84]. We will use \$3.25 as the target hourly compensation rate for the purposes of these estimates.

The original COCO release contains several different types of annotations for all 2.5 million objects in the dataset: category labels for an entire image, instance spotting for individual objects, and pixel-level instance segmentation. We will consider only the first two tasks, since they are also represented in FathomNet, and each image was observed by 5 independent annotators to ensure high recall. Lin et al.[23] estimated that the initial release of COCO required over 70,000 Turker hours to annotate 328,000 images across all three tasks, with category labels and instance spotting requiring 17,751 and 8,417 worker hours, respectively. At the stated hourly annotator rate, COCO cost about \$85,046 to annotate on AMT (Table S1).

The ImageNet2012 competition release[58] contains 1.4 million annotated images across the training, validation, and test splits for their category labeling task. The rates in Lin et al.[23] suggest that AMT workers were able to affix category labels at a rate of 92.4 images per hour. If each image in ImageNet2012 is observed once, then the task took 15,151 worker hours to complete. Like COCO, the ImageNet authors had multiple AMT workers observe each image but used a dynamic annotation scheme[22] to minimize redundant effort on easily distinguishable objects. The number of AMT workers looking at each image is thus variable and the authors do not analyze performance on a class-by-class basis. Assuming each category required an average of five workers, labeling ImageNet2012 required ∼75,755 AMT hours at a total cost of \$246,203 (Table S1). It is worth noting that these estimates do not include the costs for image generation, intellectual labor on the part of the managers, hosting fees, or compute costs for web scraping.

Fine-grained, taxonomically correct annotation is difficult to crowd-source on AMT[57]. The initial release of FathomNet annotations thus rely on domain expert annotations from the institutions generating the images. The annotation cost for MBARI's Video Lab for one technician is \$80 per hour. Expert annotators require approximately 6 months of training before achieving expert status, and the annotator will continue to learn taxonomies and animal morphology on the job. The bounding boxes for FathomNet require different amounts of time in different marine environments; midwater images typically have fewer targets, while benthic images can be very dense. Based on the Video Lab's initial annotation efforts, an experienced annotator can label ∼ 80 midwater images per hour for a $1 per image cost. The same domain experts were able to label ∼ 20 benthic images per hour or about $3 per image. The 66,039 images in the initial upload to FathomNet from MBARI are approximately evenly split between the two habitats, costing ∼$165,100 to generate the annotations (Table S1). At this hourly rate, COCO and ImageNet would cost ∼$2.1M and ∼$6.1M to annotate, respectively. We believe these costs are in-line with other annotated ocean image datasets. True domain expertise is expensive and reflects the value of an individual's training and contribution to a project. In addition to the intellectual costs of generating FathomNet, ocean data collection often requires extensive instrument development and many days of expensive ship time. To date, FathomNet largely draws from MBARI's VARS database, which is comprised of 6,190 ROV dives and represents ∼$143.7M worth of ship time. Including these additional costs underscores the value of FathomNet, especially to groups in the ocean community that are early in their data collection process.

## Acknowledgements

Seed funding for FathomNet was provided by National Geographic Society (518018 to KK), National Oceanic and Atmospheric Administration (NA18OAR4170105 to KCB), and the Monterey Bay Aquarium Research Institute through generous support from the David and Lucile Packard Foundation (to KK). Additional funding support has been provided by National Geographic Society (NGS-86951T-21 to KCB), the National Science Foundation (OTIC 1812535 and Convergence Accelerator 2137977; to KK), and the Monterey Bay Aquarium Research Institute (to KK). Additional individuals whose contributions enriched FathomNet include members of MBARI's Video Lab (Nancy Jacobsen Stout, Kyra Schlining, Susan von Thun, Kristine Walz, Larissa Lemon), Bioinspiration Lab (Joost Daniels, Paul Roberts, Krish Mehta), and Alexandra Lapides. National Geographic Society contributions were facilitated by Denley Delaney and Alan Turchik.

## Author contributions statement

KK, KCB, and BW conceived FathomNet. EO generated database statistics with early contributions from OB; EO, BW, and KK worked on representative use cases. BS designed and built the database, VARS-to-FathomNet data pipelines, API, and website back-end; KB developed the Python API and worked on data ingestion from VARS to FathomNet. EB contributed to the refinement of the FathomNet website front-end. MC facilitated data contributions and ran dataset statistics on NOAA-OER's contributions. BW developed Tator-to-FathomNet data pipelines, which included NOAA and NGS data. LL and GS generated most of the labeled data from VARS that are contained in FathomNet, and LL conducted labeling experiments for the included use cases. KK wrote the manuscript with significant contributions from EO. All authors reviewed the manuscript.

## Additional information

**Accession codes** All code and data used for this manuscript can be found on the FathomNet Code Repository[63] at `www.github.com/fathomnet` and the FathomNet database[34] at `www.fathomnet.org`. The referenced machine learning models for the benthic and midwater use cases can be found either listed in the FathomNet Model Zoo[64] at `www.github.com/fathomnet/models`, or at[44] and[47], respectively. **Competing interests** The authors declare no competing interests. Supplementary information is available for this paper, which includes video.

## Tables and Figures

**Table 1.** Data entry categories for Collections (corresponding to a single upload) and Images that make up the collections, divided into required, recommended, and suggested fields. The data entry categories for Collections are from the Darwin CORE Archive data standard.

| | Data Entry Categories | | |
|---|---|---|---|
| | Required | Recommended | Suggested |
| Collections | • Owner's institution<br>• Rights holder (use owner's institution if not specified)<br>• Contributor's email<br>• Record type (images)<br>• Modified field (upload date)<br>• UUID<br>• URL<br>• Data format (CSV+) | • Bibliographic citation<br>• Access rights (for more generous use)<br>• Basis of record (human or machine)<br>• dataset language | • Collection code<br>• Collection ID<br>• dataset generalizations<br>• dataset name<br>• Dynamic properties<br>• Information withheld<br>• Institution code<br>• Institution ID<br>• References |
| Images | • Image URL<br>• Bounding box coordinates (x, y, width, height)<br>• Concept name | • Latitude/Longitude<br>• Depth<br>• Timestamp (ISO8601)<br>• Imaging type<br>• Observer<br>• Altitude | • Group of<br>• Occluded<br>• Truncated<br>• AltConcept |

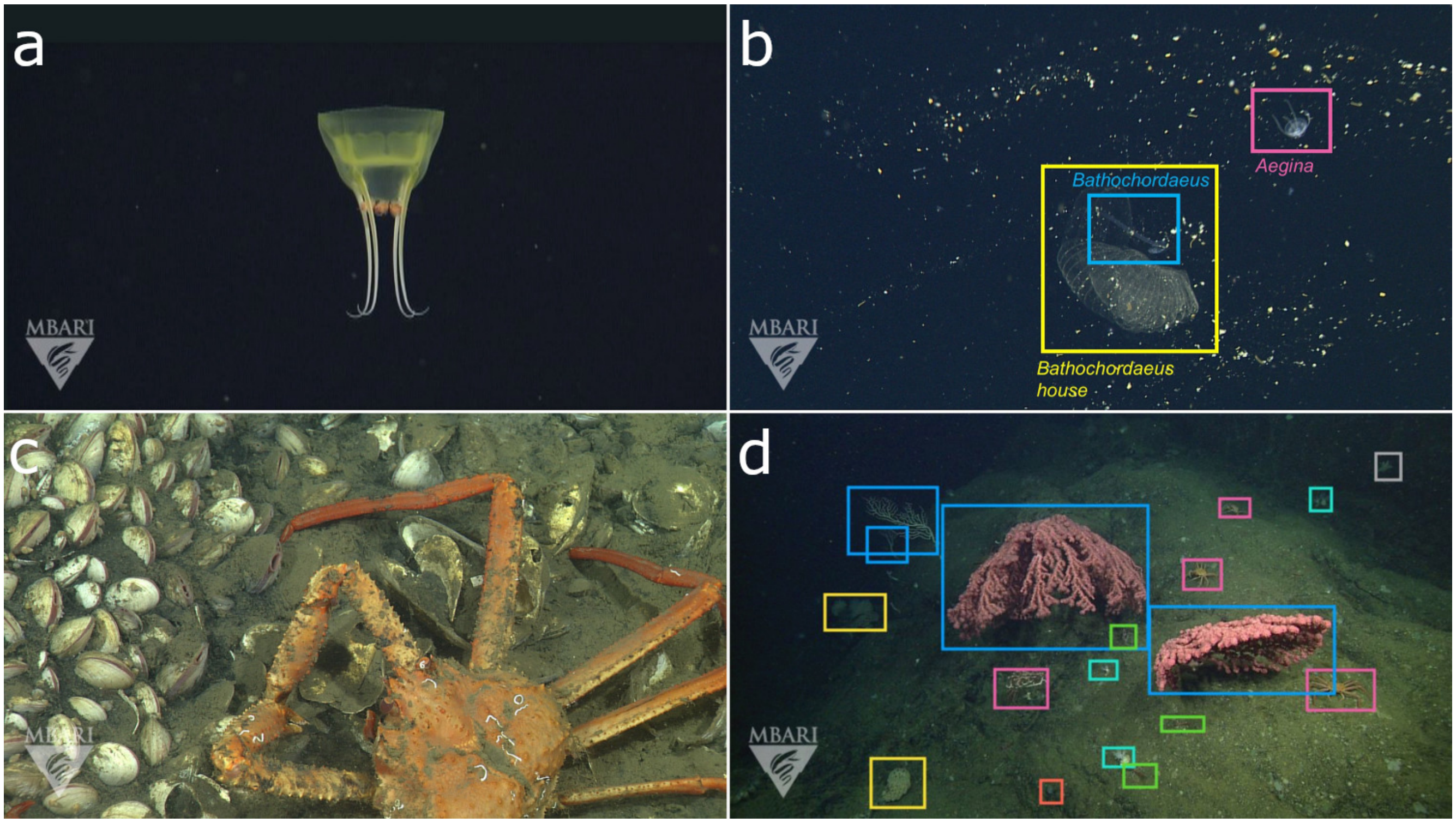

**Figure 1. Labeled image data for deep learning algorithms require an annotation and localization for a single concept.** Images showing the disparity between (a,c) traditionally annotated visual data in MBARI's VARS database and (b,d) the level of annotations and localizations (or bounding boxes) that are required for the data science community. Top (a,b) and bottom (c,d) rows show images for midwater and benthic environments, respectively. (a,b) and (c,d) are example images from VARS that have been assigned a single-concept annotation for the genuses *Aegina* (b, pink) and *Chionoecetes* (d, pink), respectively. Although images in the left column (a,c) are largely iconic views of a single-concept, images in the right column (b,d) show non-iconic views of multiple concepts that additionally include (b) *Bathochordaeus* (blue), *Bathochordaeus* house (yellow), (d) *Paragorgia* (blue), *Asteroidea* (pink), *Psolus* (cyan), *Pandalopsis* (green), *Heterochone* (yellow), and *Tunicata* (red). Label text has been removed for clarity in d.

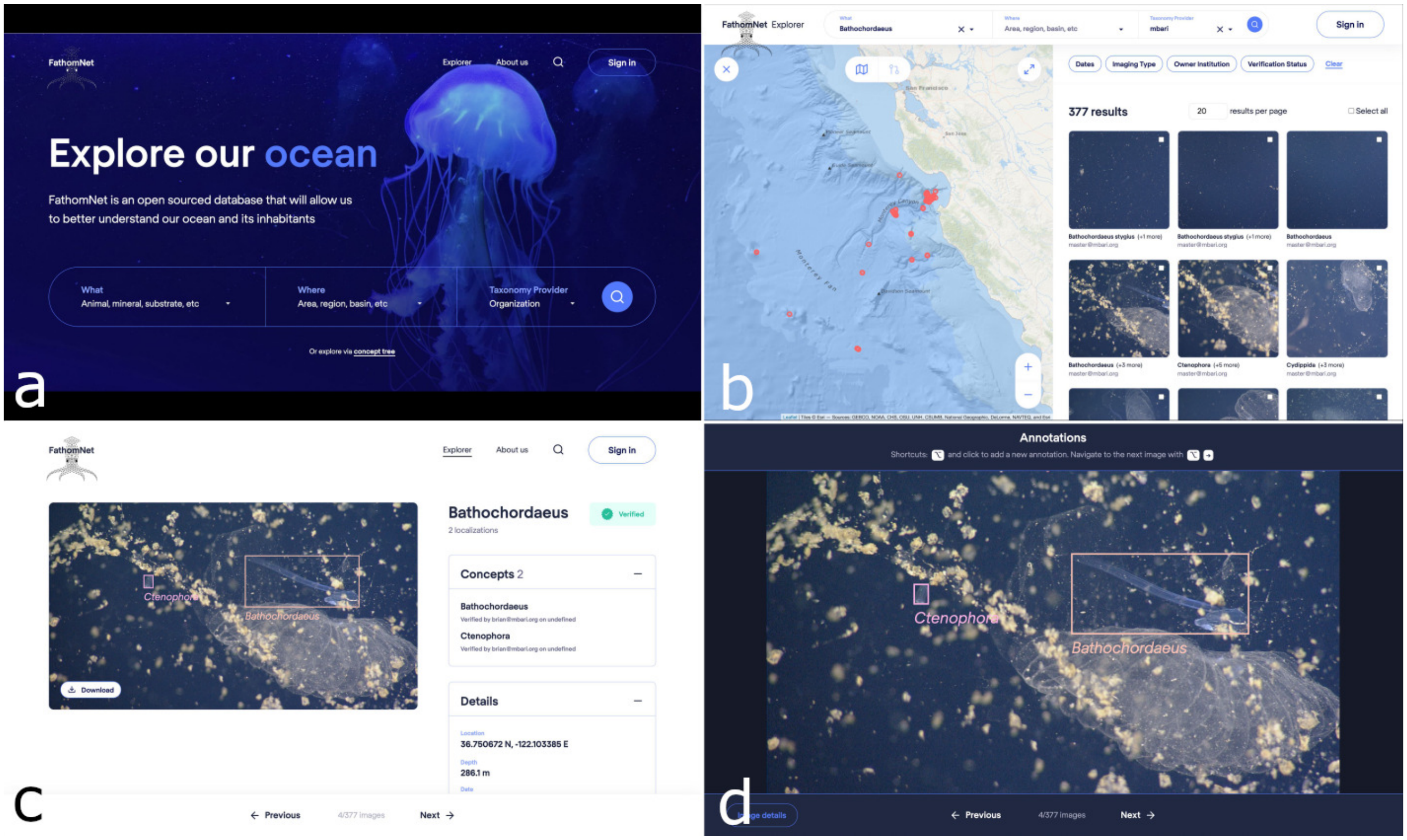

**Figure 2. FathomNet data portal can be accessed at www.fathomnet.org.** The website contains features that include (a) a simple search bar for terms in the concept tree, (b) filtered searches where images can be displayed based on geographic location or terms within the concept tree (among others), (c) image display pages where concepts, details, and contributors' information is shown, and (d) basic annotation and localization tool to allow users to augment or correct uploaded data in the database.

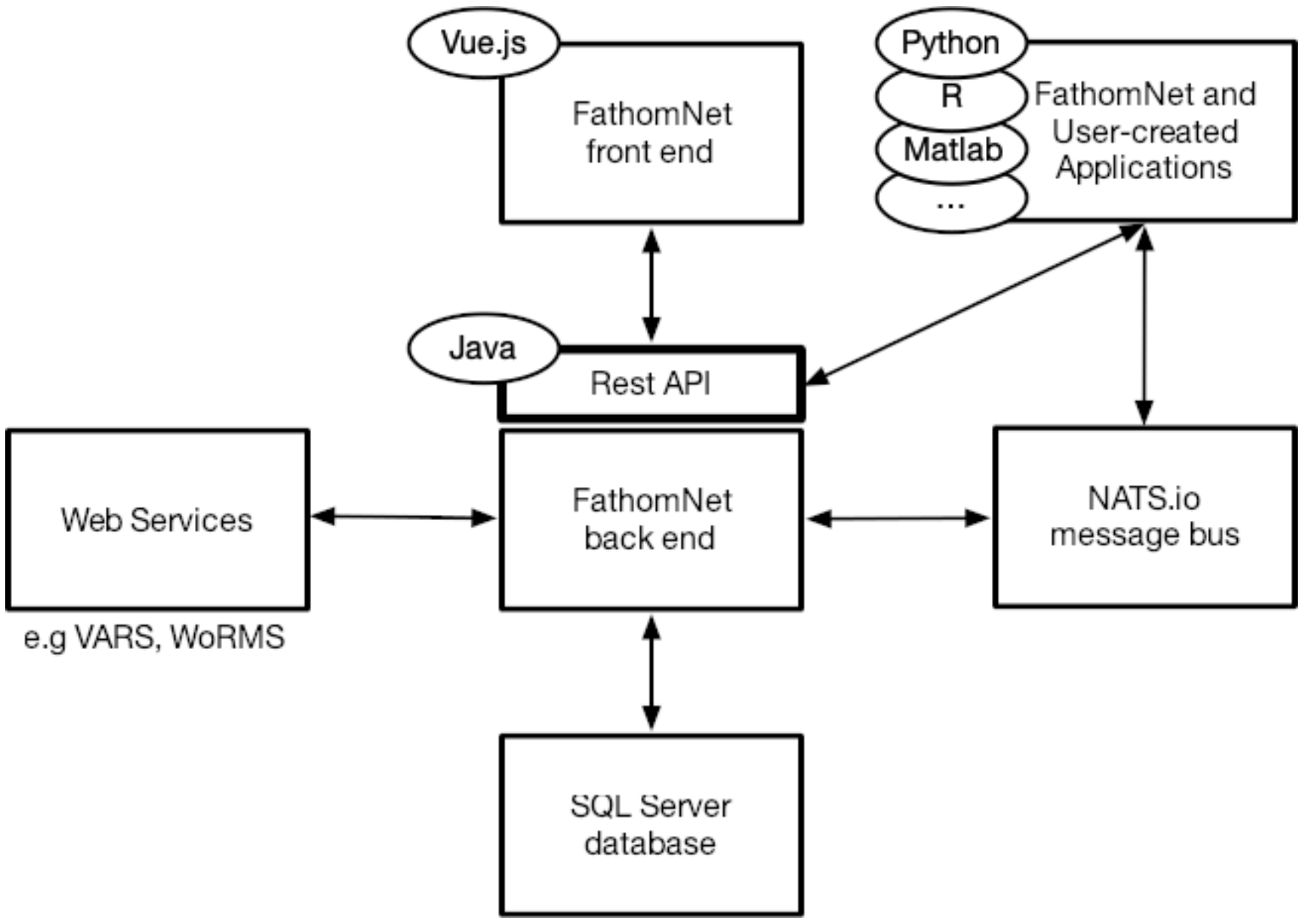

**Figure 3. FathomNet database software architecture.** The FathomNet software architecture includes a Javascript Rest API, SQL Server database, a NATS.io messaging bus, and external web services (e.g., WoRMS, VARS, MarineRegions).

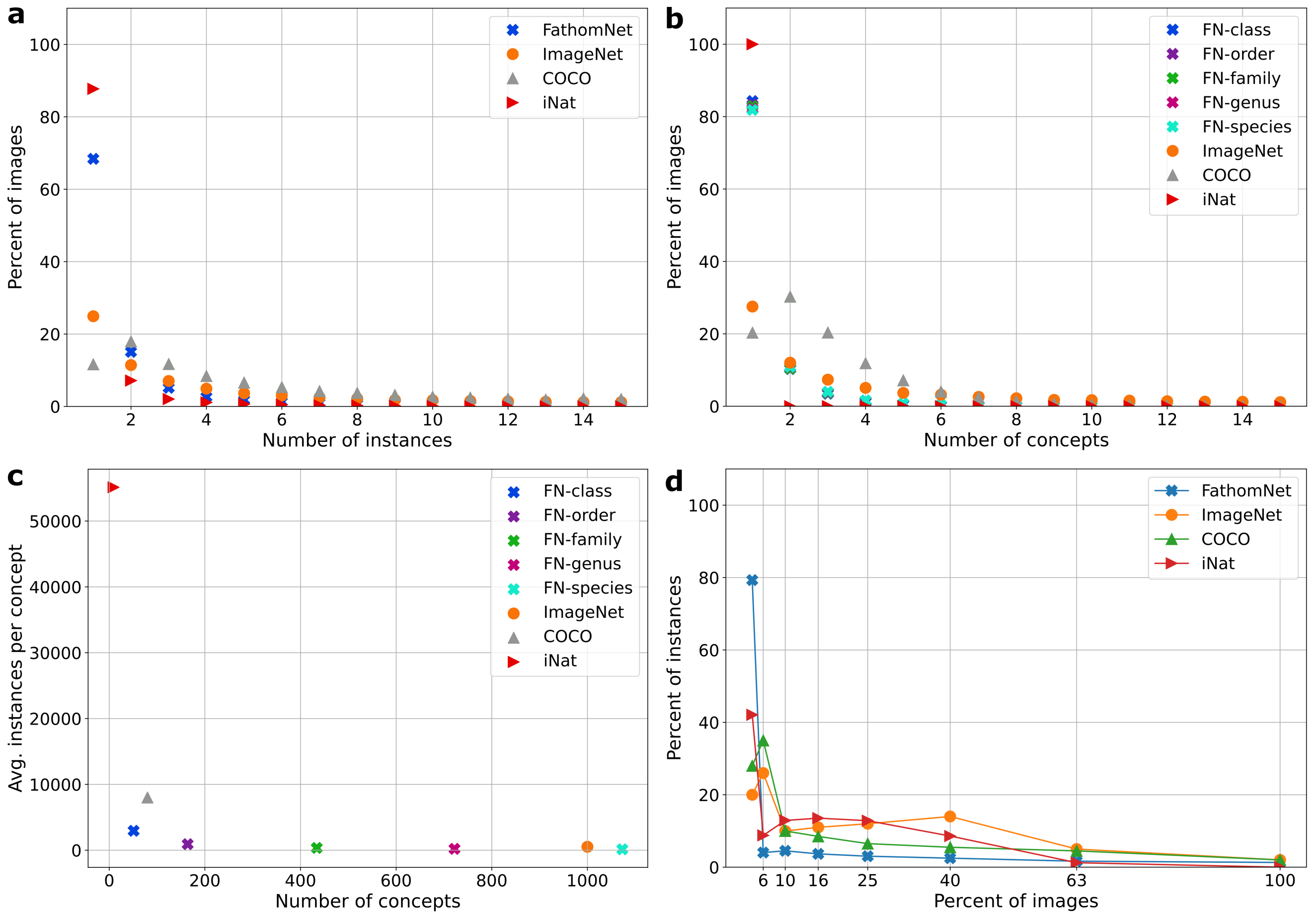

**Figure 4.** Dataset statistics contextualizing FathomNet relative to COCO, ImageNet, and iNat2017. Statistics for COCO were computed from the 2014 training split, ImageNet numbers were generated from the 2015 localization data split, and iNat2017 metrics were derived from the 2017 training data. (a) Percent of images containing a number of instances, or localizations. (b) Percent of images displaying a given number of concepts. FathomNet annotations are displayed broken down at different levels of taxonomic rank. (c) The average number of instances versus the average number of concepts over the entirety of each dataset. (d) The distribution of relative instance size. Most objects in FathomNet and iNat are small as compared to ImageNet and COCO.

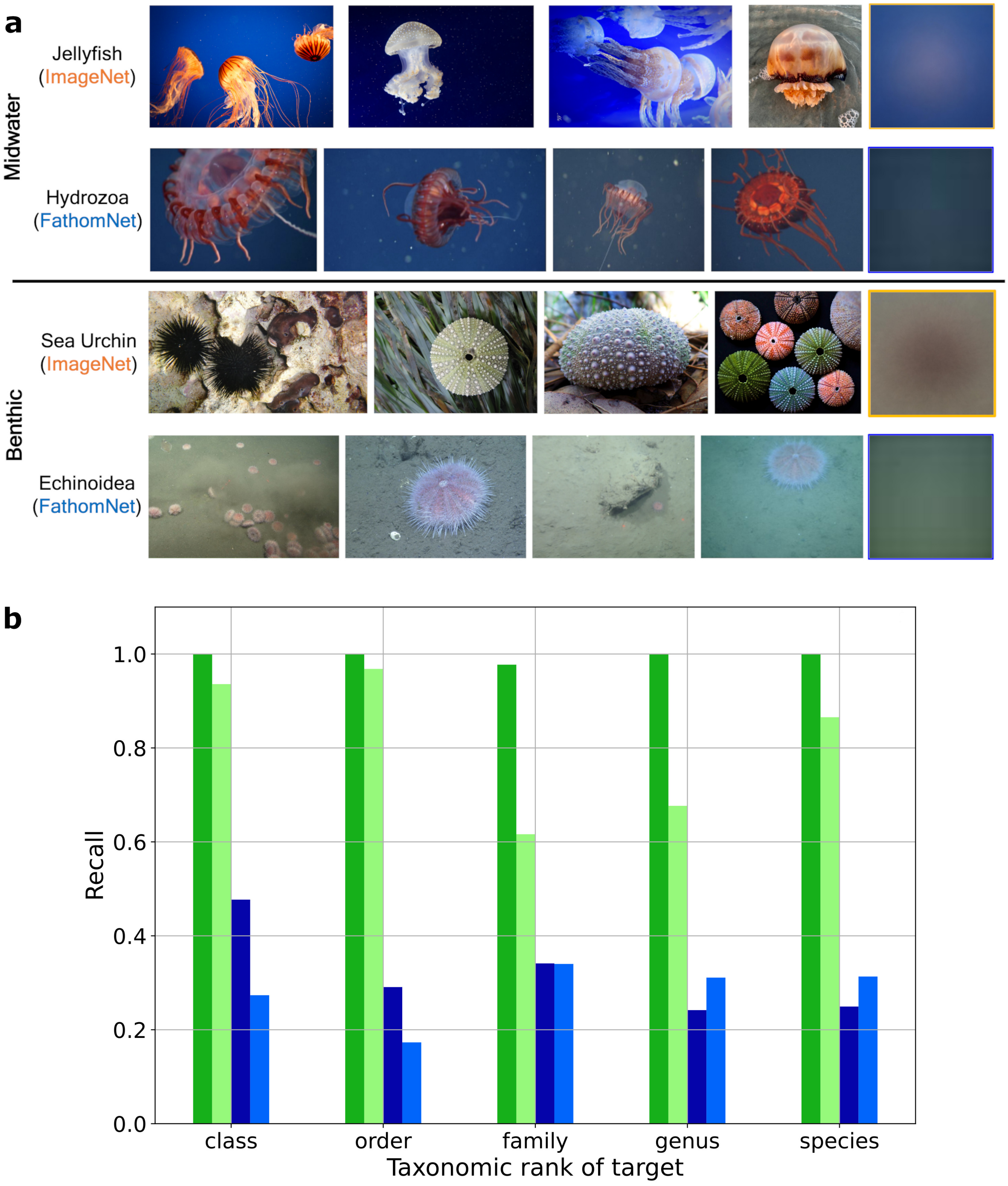

**Figure 5.** Dataset visualizations contextualizing FathomNet relative to ImageNet. (a) The average pixel values of images in ImageNet and FathomNet. Images from a midwater and benthic concept in ImageNet (hosted on Flickr) were approximately matched to concepts available in FathomNet, and an equal subset of images from both repositories were resized and averaged pixel-wise (far-right column). The uniform image patches from FathomNet suggest that there is more diversity in poses and object sizes relative to labeled data from ImageNet. FathomNet image credits: MBARI. (b) Concept coverage at different taxonomic ranks within FathomNet. 50 images were selected randomly at each level of the taxonomic tree and checked for annotation completeness by a human expert. Green bars correspond to the midwater concept *Bathocordaeus mcnutti* and blue bars correspond to the benthic concept *Gersemia juliepackardae*. Darker colors are the target concept and lighter colors are any other concept present in the frame. Benthic images are typically missing more annotations than those from midwater habitats.

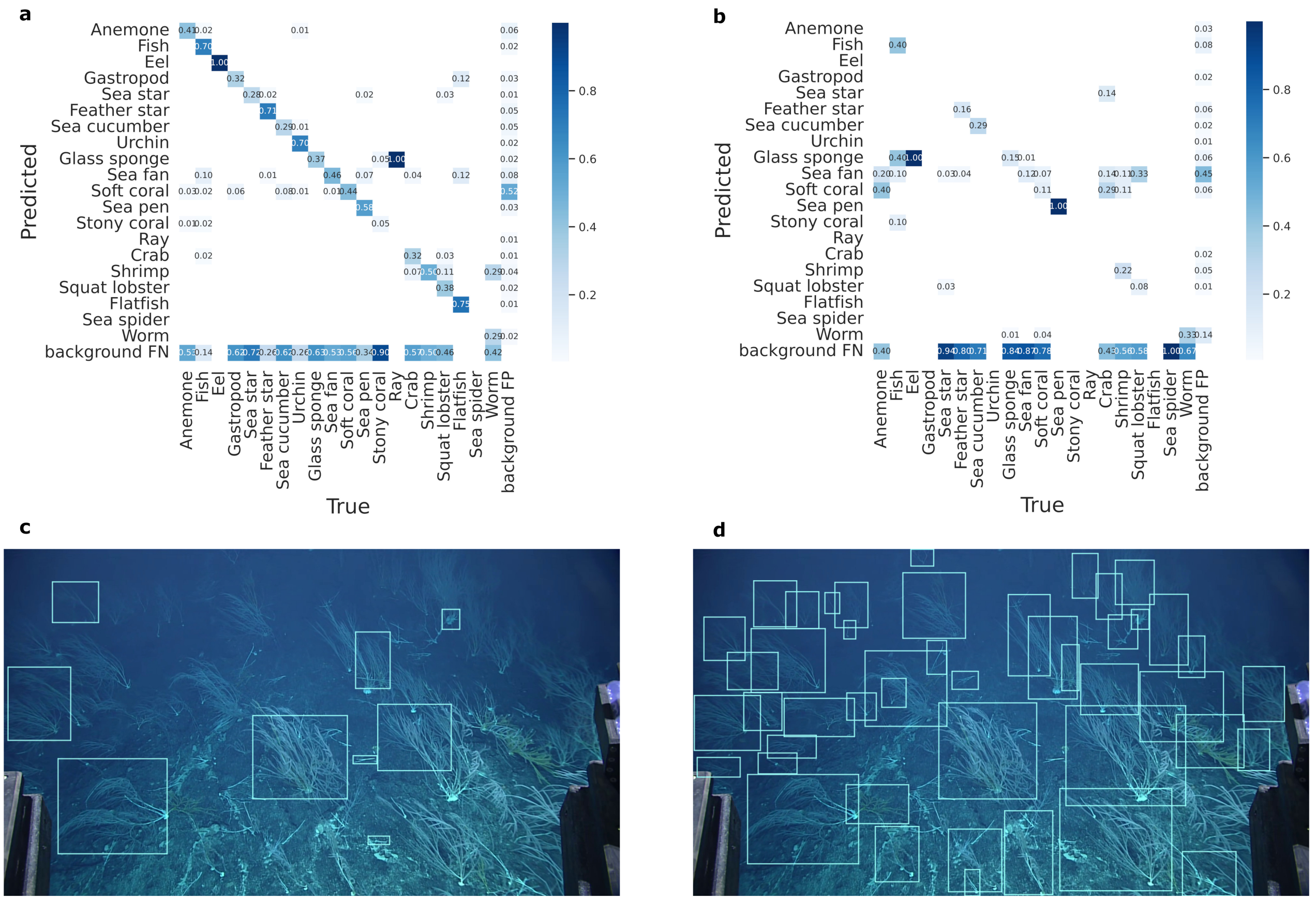

**Figure 6.** Results from the NOAA Benthic use case. (a) Validation confusion matrix from the RetinaNet object detector trained on supercategory data from Monterey Bay. The row of background false negatives is mostly due to incomplete original annotation data. (b) Confusion matrix from the model applied to NOAA data from the Musician Seamount. Note that the vast majority of the targets were sea fans. (c) Model output from a single frame of NOAA video data. (d) Ground truth annotations for the same frame. Note the density, variability in coral morphology, and overlapping individuals.

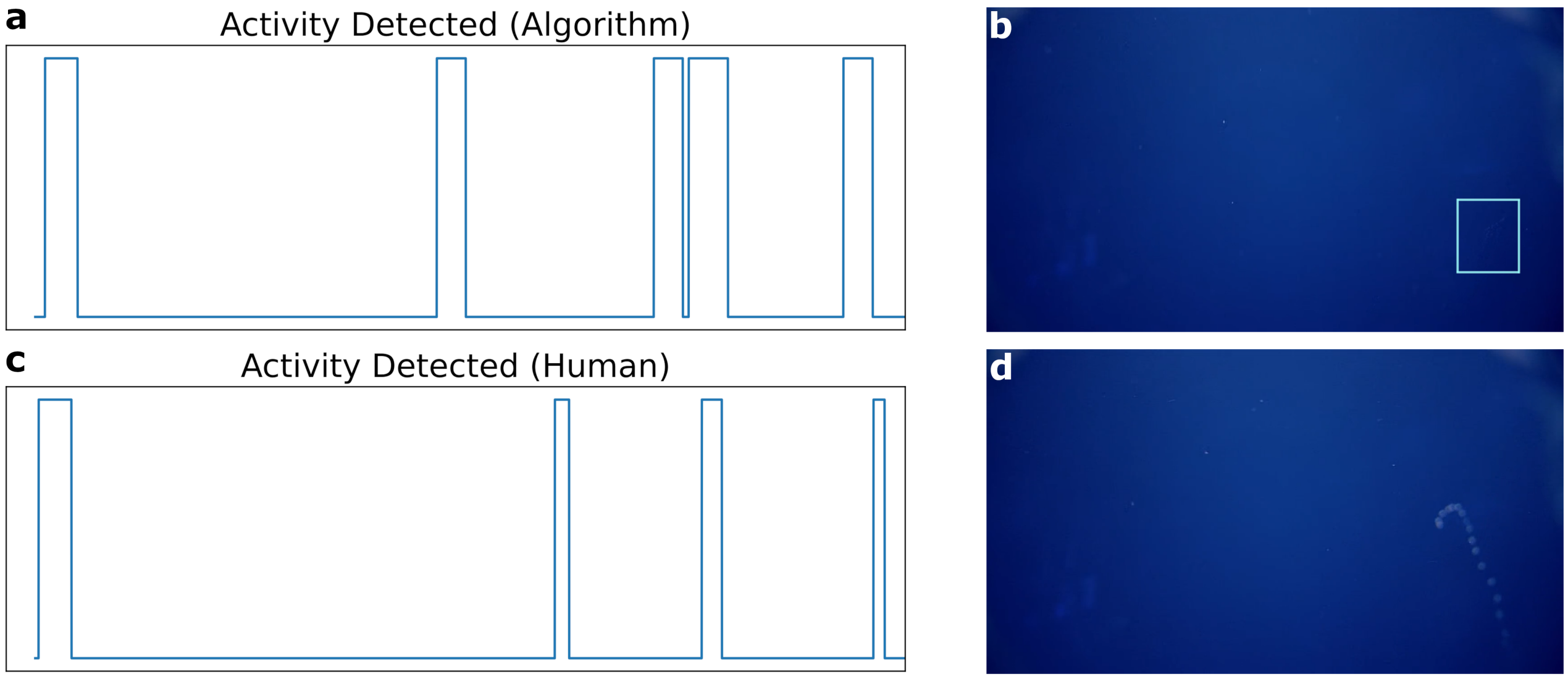

**Figure 7.** Comparison of activity recognition output from object detection (top-left) and a human expert (bottom-left) for one video collected by NOAA's ROV *Deep Discoverer* during the CAPSTONE expedition[45]. In one instance (top-right), the algorithm (indicated by a bounding box) correctly identifies an animal with low contrast (image has been contrast enhanced). In another (bottom-right), the algorithm fails to identify an animal that is blurry due to unfocused optics.